\documentclass[10pt,twocolumn,letterpaper]{article}

\usepackage[pagenumbers]{iccv} 

\usepackage{amsmath,amsfonts,bm}

\def\eqref#1{equation~\ref{#1}}

\def\1{\bm{1}}

\def\rvh{{\mathbf{h}}}

\def\rvx{{\mathbf{x}}}

\def\rvz{{\mathbf{z}}}

\DeclareMathAlphabet{\mathsfit}{\encodingdefault}{\sfdefault}{m}{sl}
\SetMathAlphabet{\mathsfit}{bold}{\encodingdefault}{\sfdefault}{bx}{n}

\newcommand{\bk}{\boldsymbol{k}}

\definecolor{iccvblue}{rgb}{0.21,0.49,0.74}
\usepackage[pagebackref,breaklinks,colorlinks,allcolors=iccvblue]{hyperref}

\usepackage{multirow}
\usepackage{marvosym}
\usepackage[accsupp]{axessibility}

\def\paperID{7784} 
\def\confName{ICCV}
\def\confYear{2025}

\title{FreeScale: Unleashing the Resolution of Diffusion Models \\ via Tuning-Free Scale Fusion}

\author{{Haonan Qiu}$^1$, 
{Shiwei Zhang}$^{2,}$\textsuperscript{\Letter},
{Yujie Wei}$^3$,
{Ruihang Chu}$^2$, 
{Hangjie Yuan}$^2$, \\
{Xiang Wang}$^2$, 
{Yingya Zhang}$^2$, 
{Ziwei Liu}$^{1,}$\textsuperscript{\Letter}\\
\small $^{1}$Nanyang Technological University \qquad
\small $^{2}$Alibaba Group \qquad
\small $^{3}$Fudan University \\
\\
Project Page: \href{http://haonanqiu.com/projects/FreeScale.html}{http://haonanqiu.com/projects/FreeScale.html}
}

\begin{document}

\twocolumn[{
\maketitle
    \centering
    \vspace{-2.4em}
    \includegraphics[width=0.99\textwidth]{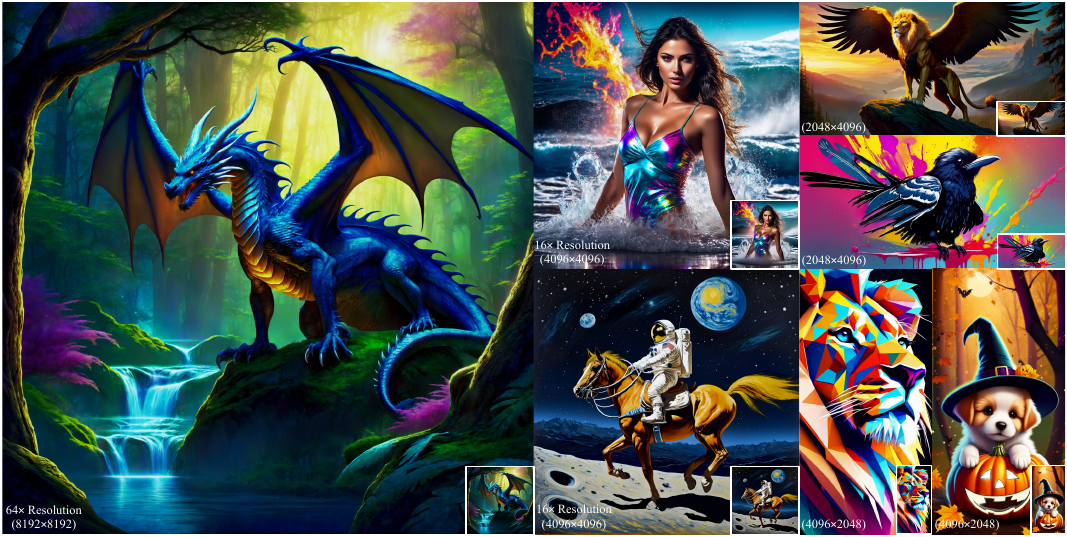}
    \vspace{-0.8em}
    \captionof{figure}{
        \textbf{Gallery of FreeScale.} Original SDXL~\cite{sdxl} can only generate images with a resolution of up to $1024^2$ without losing quality, while FreeScale successfully extends SDXL to generate $8192^2$ images without any fine-tuning. All generated images are produced using a single A800 GPU. Best viewed \textbf{ZOOMED-IN}.
    }
    \label{first_figure}
    \vspace{1.0em}
    }
]

\begin{abstract}

Visual diffusion models achieve remarkable progress, yet they are typically trained at limited resolutions due to the lack of high-resolution data and constrained computation resources, hampering their ability to generate high-fidelity images or videos at higher resolutions. 
Recent efforts have explored tuning-free strategies to exhibit the untapped potential higher-resolution visual generation of pre-trained models.
However, these methods are still prone to producing low-quality visual content with repetitive patterns.
The key obstacle lies in the inevitable increase in high-frequency information when the model generates visual content exceeding its training resolution, leading to undesirable repetitive patterns deriving from the accumulated errors.
To tackle this challenge, we propose \textbf{FreeScale}, a tuning-free inference paradigm to enable higher-resolution visual generation via scale fusion. Specifically, FreeScale processes information from different receptive scales and then fuses it by extracting desired frequency components. 
Extensive experiments validate the superiority of our paradigm in extending the capabilities of higher-resolution visual generation for both image and video models. 
Notably, compared with previous best-performing methods, FreeScale unlocks the \textbf{8k}-resolution text-to-image generation for the first time.

\end{abstract}
\section{Introduction}
\label{sec:introduction}

Diffusion models have revolutionized visual generation~\cite{sdxl,pixart-alpha,wang2023modelscope,chen2024videocrafter2,yang2024cogvideox,zhang2023adding}, empowering individuals without any artistic expertise to effortlessly create distinctive and personalized designs, graphics, and short films using specific textual descriptions. 
Nonetheless, current visual diffusion models are generally trained on data with limited resolution, such as $512^2$ for SD 1.5~\cite{ldm}, $1024^2$ for SDXL~\cite{sdxl}, and $320\times512$ for VideoCrafter2~\cite{chen2024videocrafter2}, hampering their ability to generate high-fidelity images or videos at higher resolutions.
Given the scarcity of high-resolution visual data and the substantially greater model capacity required for modeling such data, recent efforts have focused on employing tuning-free strategies for high-resolution visual generation to inherit the strong generation capacities of existing pre-trained diffusion models. 

Despite the advances achieved by existing methods, they are still prone to producing low-quality images or videos, particularly manifesting as repetitive object occurrences and unreasonable object structures.
ScaleCrafter~\cite{he2023scalecrafter} puts forward that the primary cause of the object repetition issue is the limited convolutional receptive field and uses dilated convolutional layers to achieve tuning-free higher-resolution sampling. But the generated results of ScaleCrafter still suffer from the problem of local repetition. Inspired by MultiDiffusion~\cite{bar2023multidiffusion} fusing the local patches of the whole images, DemoFusion~\cite{du2024demofusion} designed a mechanism by fusing the local patches and global patches, almost eliminating the local repetition. Essentially, this solution just transfers the extra signal of the object to the background, leading to small object repetition generation. FouriScale~\cite{huang2024fouriscale} reduces those extra signals by removing the high-frequency signals of the latent before the convolution operation. Although FouriScale completely eliminates all types of repetition, the generated results always have weird colors and textures due to its violent editing on the frequency domain. 

To generate satisfactory visual contents without any unexpected repetition, we propose \textbf{FreeScale}, a tuning-free inference paradigm that enables pre-trained image and video diffusion models to generate vivid higher-resolution results. 
Building on past effective modules~\cite{guo2024make,he2023scalecrafter}, we first propose tailored self-cascade upscaling and restrained dilated convolution to gain the basic visual structure and maintain the quality in higher-resolution generation.
To further eliminate all kinds of unexpected object repetitions, FreeScale processes information from different receptive scales and then fuses it by extracting desired frequency components, ensuring both the structure's overall rationality and the object's local quality. 
This fusion is smoothly integrated into the original self-attention layers, thereby bringing only minimal additional time overhead. 
Finally, we demonstrate the effectiveness of our model on both the text-to-image model and the text-to-video model, pushing the boundaries of image generation even up to an 8k resolution.

Our contributions are summarized as follows:

\begin{itemize}

\item We propose \textbf{FreeScale}, a tuning-free inference paradigm to enable pre-trained diffusion models to generate vivid higher-resolution results via fusing the information from different scales.
\item We empirically evaluate our approach on both the text-to-image model and the text-to-video model, demonstrating the effectiveness of our model.
\item Compared to other state-of-the-art tuning-free methods, we unlock the \textbf{8k}-resolution text-to-image generation for the first time. 

\end{itemize}

\section{Related Work}
\label{sec:related}

\noindent\textbf{Diffusion Models for Visual Generation.}
The advent of diffusion models has transformed the landscape of image and video generation by enabling the production of exceptionally high-quality outputs~\cite{sdxl,pixart-alpha,wang2023modelscope,chen2024videocrafter2,yang2024cogvideox,zhang2023adding,yuan2023instructvideo,si2023freeu,wei2024dreamvideo}. Initial breakthroughs like DDPM~\cite{ddpm} and Guided Diffusion~\cite{guided-diffusion} demonstrated that diffusion processes could yield remarkable image quality. To enhance computational efficiency, LDM~\cite{ldm} introduced latent space diffusion, which operates in a compressed space, significantly lowering the computational burden and training demands; this method laid the groundwork for Stable Diffusion. Building on this, SDXL~\cite{sdxl} further advanced high-resolution image synthesis. Inspired by DiT~\cite{peebles2023scalable}, Pixart-alpha~\cite{pixart-alpha} adopted a transformer-based architecture, achieving both high fidelity and cost-effective image generation. 

For video generation, VDM~\cite{vdm} pioneered the application of diffusion in this domain, followed by LVDM~\cite{lvdm}, which extended the method to propose a hierarchical latent video diffusion framework capable of generating extended video sequences. To bridge text-to-image and text-to-video (T2V) capabilities, Align-Your-Latents~\cite{blattmann2023align} and AnimateDiff~\cite{guo2023animatediff} introduced temporal transformers into existing T2I models. VideoComposer~\cite{videocomposer} then offered a controllable T2V generation approach, allowing precise management of spatial and temporal cues. VideoCrafter~\cite{chen2023videocrafter1, chen2024videocrafter2} and SVD~\cite{svd} scaled these latent video diffusion models to handle extensive datasets. Lumiere~\cite{lumiere} proposed temporal downsampling within a space-time U-Net for greater efficiency. 
Finally, CogVideoX~\cite{yang2024cogvideox} and Pyramid Flow~\cite{jin2024pyramidal} two recent highly regarded open-source models, showcase impressive video generation capabilities, demonstrating the superior performance of DiT structure in video generation.

Since the DiT structure models often take up more memory, achieving high-resolution generation on a single GPU is difficult even in the inference phase. Therefore, we still use the U-Net structure models in this work. We chose SDXL~\cite{sdxl} as our pre-trained image model, and VideoCrafter2~\cite{chen2024videocrafter2} as our pre-trained video model.

\begin{figure*}[t]
\centering
\includegraphics[width=0.99\linewidth]{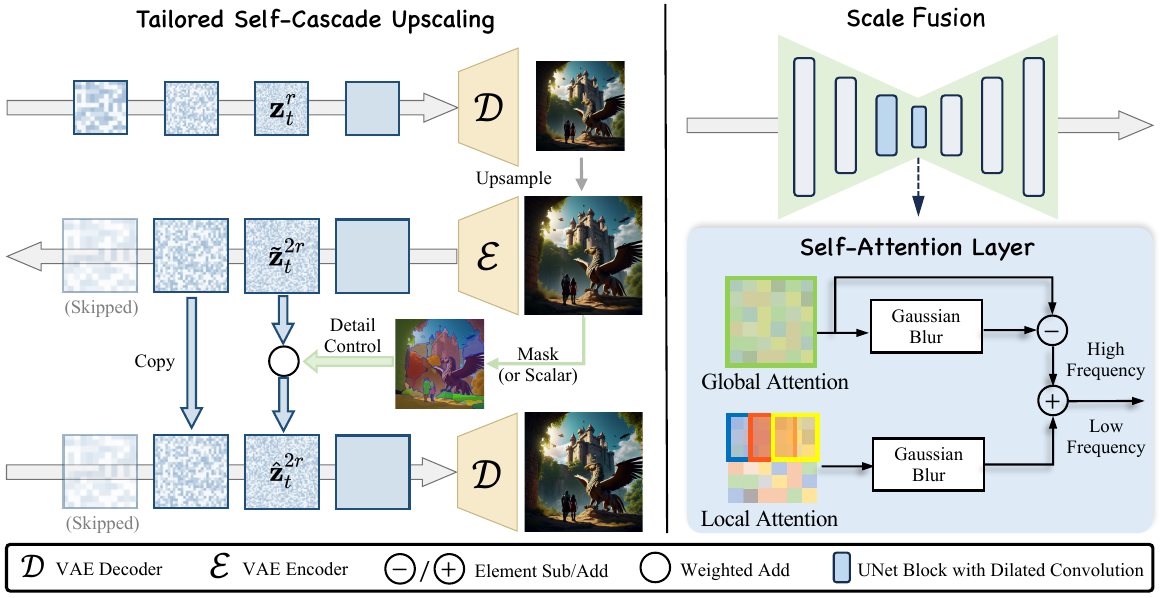}\vspace{-1.0em}
\caption{
\textbf{Overall framework of FreeScale.} 
(a) \textit{Tailored Self-Cascade Upscaling}. FreeScale starts with pure Gaussian noise and progressively denoises it using the training resolution. 
An image is then generated via the VAE decoder, followed by upscaling to obtain a higher-resolution one.
We gradually add noise to the latent of this higher-resolution image and incorporate this forward noise into the denoising process of the higher-resolution latent with the use of \textit{restrained dilated convolution}. 
Additionally, for intermediate latent steps, we enhance high-frequency details by applying region-aware detail control using masks derived from the image.
(b) \textit{Scale Fusion}. 
During denoising, we adapt the self-attention layer to a global and local attention structure. 
By utilizing Gaussian blur, we fuse high-frequency details from global attention and low-frequency semantics from local attention, serving as the final output of the self-attention layer.
}
\vspace{-1.0em}
\label{fig:framework}
\end{figure*}

\noindent\textbf{Higher-Resolution Visual Generation.}
High-resolution visual synthesis is a classic challenge in the generative field due to the difficulty of collecting plenty of high-resolution data and the requirement of substantial computational resources.
Recent methods for higher-resolution generation can mainly be divided into two categories: \text{1)} training/tuning methods with high-resolution data and large models~\cite{teng2023relay,hoogeboom2023simple,ren2024ultrapixel,liu2024linfusion,guo2024make, zheng2024any, cheng2024resadapter}, or \text{2)} tuning-free methods without any additional data requirement~\cite{haji2024elasticdiffusion, lin2024cutdiffusion, lee2023syncdiffusion, jin2023training, hwang2024upsample, cao2024ap, zhang2024hidiffusion, kim2024diffusehigh}. 
Training with high-resolution data on larger models should be a more fundamental solution. However, high-resolution visual data only accounts for a small proportion. Meanwhile, targeting for modeling higher-resolution data demands a notably increased requirement in model capacity. Based on current data and calculation resources, tuning-free approaches are more achievable for high-resolution generation.

One straightforward approach is to generate visual patches of the same resolution as the training data and then stitch them together. Although eliminating the training-inference gap, this method results in disconnected and incoherent patches. MultiDiffusion~\cite{bar2023multidiffusion} addresses this problem by fusing patches smoothly during the denoising process. DemoFusion~\cite{du2024demofusion} utilizes this mechanism and adds global perception to ensure the rationality of the overall layout. However, this solution easily leads to the generation of small object repetition. ScaleCrafter~\cite{he2023scalecrafter} argues that the object repetition issue is mainly caused by the limited convolutional receptive field and uses dilated convolutional layers to enlarge the convolutional receptive field. Although successful in removing small object repetition, ScaleCrafter suffers from a new problem of local repetition. FouriScale~\cite{huang2024fouriscale} concludes that all types of repetitions are from the non-alignment of frequency domain on different scales. FouriScale removes the high-frequency signals of the latent prior to convolution operation and achieves no repetition at all. But this violent editing operation on the frequency domain leads to strange results with unnatural colors and textures. Another solution is directly removing the text semantics from unexpected areas in the input level~\cite{lin2024accdiffusion,liu2024hiprompt}. However, it only works for small object repetition and will suffer information leakage through the temporal layers in the video generation.
With the additional pose as input, BeyondScene~\cite{kim2024beyondscene} has achieved 8k human image generation. However, its scope is limited to human image generation due to the requirement of additional pose input. FreeScale is the first 8k-resolution text-to-image generation method without these constraints.

\section{Methodology}
\label{sec:methodology}

\subsection{Preliminaries}
\label{subsec:preliminary}

\noindent\textbf{Latent Diffusion Models (LDM)} first encodes a given image $\rvx$ to the latent space $\rvz$ via the encoder of the pre-trained auto-encoder $\mathcal{E}$: $z=\mathcal{E}(x)$. Then a forward diffusion process is used to gradually add noise to the latent data $\rvz_0 \sim p(\rvz_0)$ and learn a denoising model to reverse this process. The forward process contains $T$ timesteps, which gradually add noise to the latent sample $\rvz_0$ to yield $\rvz_t$ through a parameterization trick:
\begin{equation}
\begin{aligned}
 &q(\rvz_t|\rvz_{t-1}) = \mathcal{N}(\rvz_t;\sqrt{1-\beta_t}\rvz_{t-1},\beta_t \mathbf{I}), \\
 &q(\rvz_t|\rvz_0) = \mathcal{N}(\rvz_t; \sqrt{\bar\alpha_t}\rvz_0, (1-\bar\alpha_t)\mathbf{I}),
\end{aligned}
\end{equation}
where $\beta_t$ is a predefined variance schedule, $t$ is the timestep, $\bar\alpha_t = \prod_{i=1}^t \alpha_i$, and $\alpha_t = 1-\beta_t$.
The reverse denoising process obtains less noisy latent $\rvz_{t-1}$ from the noisy input $\rvz_t$ at each timestep:
\begin{equation}
p_\theta\left(\boldsymbol{x}_{t-1} \mid \boldsymbol{x}_t\right)=\mathcal{N}\left(\boldsymbol{x}_{t-1} ; \boldsymbol{\mu}_\theta\left(\rvz_t, t\right), \boldsymbol{\Sigma}_\theta\left(\rvz_t, t\right)\right),
\end{equation}
where $\boldsymbol{\mu}_\theta$ and $\boldsymbol{\Sigma}_\theta$ are determined through a noise prediction network $\boldsymbol{\epsilon}_{\theta}\left(\rvz_t, t\right)$ with learnable parameters $\theta$. 

\subsection{Tailored Self-Cascade Upscaling}

Directly generating higher-resolution results will easily produce several repetitive objects, losing the reasonable visual structure that was originally good. To address this issue, we utilize a self-cascade upscaling framework from previous works~\cite{du2024demofusion,guo2024make}, which progressively increases the resolution of generated results:
\begin{equation}
\tilde{\rvz}_K^{2r} \sim \mathcal{N}\left(\sqrt{\bar{\alpha}_K} \phi\left(\rvz_0^{r}\right), \sqrt{1-\bar{\alpha}_K} \mathbf{I}\right),
\label{eq:cascade}
\end{equation}
where $\tilde{\rvz}$ means the noised intermediate latent, $r$ is the resolution level ($1$ represents original resolution, $2$ represents the twice height and width), and $\phi$ is an upsampling operation.
Specifically, FreeScale will denoise using the training resolution. The intermediate results will then be gradually up-sampled. In the higher resolution, blurry details from the upsampling will be removed by adding noise (to the level of timestep $K$) and denoising.
In this way, the framework will generate a reasonable visual structure in low resolution and maintain the structure when generating higher-resolution results.

There are two options for $\phi$: directly upsampling in latent ($\phi\left(\rvz\right) = \text{UP}(\rvz)$) or upsampling in RGB space ($\phi\left(\rvz\right) = \mathcal{E}(\text{UP}(\mathcal{D}(\rvz)))$, where $\mathcal{E}$ and $\mathcal{D}$ are the encoder and decoder of pre-trained VAE, respectively.
Upsampling in RGB space is closer to human expectations but will add some blurs. We empirically observe that these blurs will hurt the video generation but help to suppress redundant over-frequency information in the image generation. Therefore, we adopt upsampling in RGB space for higher-solution image generation and latent space upsampling in higher-solution video generation.

\noindent\textbf{Flexible Control for Detail Level.} Different from super-resolution tasks, FreeScale will endlessly add more details as the resolution grows. This behavior will hurt the generation when all reasonable details are generated. To control the level of newly generated details, we modify $p_\theta\left(\mathbf{z}_{t-1} \mid \mathbf{z}_t\right)$ to $p_\theta\left(\mathbf{z}_{t-1} \mid \hat{\mathbf{z}}_t\right)$ with:
\begin{equation}
\hat{\mathbf{z}}_t^{r}=c \times \tilde{\mathbf{z}}_t^{r}+\left(1-c\right) \times \mathbf{z}_t^r,
\label{eq:level}
\end{equation}
where $c=\left(\left(1+\cos \left(\frac{T-t}{T} \times \pi\right)\right) / 2\right)^{\alpha}$ is a scaled cosine decay factor with a scaling factor $\alpha$. 

Even in the same image, the detail level varies in different areas. To achieve more flexible control, $\alpha$ can be a 2D-tensor and varies spatially. In this case, users can assign different values for different semantic areas according to $\mathcal{D}\left(\rvz_0^{r}\right)$ calculated in the previous process already.

\subsection{Restrained Dilated Convolution}

ScaleCrafter~\cite{he2023scalecrafter} observes that the primary cause of the object repetition issue is the limited convolutional receptive field and proposes dilated convolution to solve it. Given a hidden feature map $\rvh$, a convolutional kernel $\bk$, and the dilation operation $\Phi_d(\cdot)$ with factor $d$, the dilated convolution can be represented as:
\begin{equation}
f_{\boldsymbol{k}}^d(\rvh)=\rvh \circledast \Phi_d(\boldsymbol{k}),\left(\rvh \circledast \Phi_d(\boldsymbol{k})\right)(o)=\sum_{s+d \cdot t=p} \rvh(p) \cdot \boldsymbol{k}(q),
\label{eq:dilation}
\end{equation}
where $o$, $p$, and $q$ are spatial locations used to index the feature or kernel. $\circledast$ denotes convolution operation.

To avoid catastrophic quality decline, ScaleCrafter~\cite{he2023scalecrafter} only applies dilated convolution to some layers of UNet while still consisting of several up-blocks. However, we find that dilated convolution in the layers of up-blocks will bring many messy textures. Therefore, unlike previous works, we only apply dilated convolution in the layers of down-blocks and mid-blocks. In addition, the last few timesteps only render the details of results and the visual structure is almost fixed. Therefore, we use the original convolution in the last few timesteps.

\begin{table*}[t]
\centering
\vspace{-2mm}
\caption{\textbf{Image quantitative comparisons with other baselines.} FreeScale achieves the best or second-best scores for all quality-related metrics with negligible additional time costs. The best results are marked in \textbf{bold}, and the second best results are marked by \underline{underline}.}
\vspace{-2mm}
\label{tab:comp_img}
\scalebox{0.82}{\begin{tabular}{@{}l|cccccc|cccccc@{}}
\toprule
\multirow{2}{*}{\bf Method} & \multicolumn{6}{c|}{$2048^2$}        & \multicolumn{6}{c}{$4096^2$}        \\ \cmidrule(l){2-13} 
                        &FID $\downarrow$ &KID $\downarrow$ &$\text{FID}_c$ $\downarrow$ & $\text{KID}_c$ $\downarrow$ & \text{IS} $\uparrow$ & Time (min) $\downarrow$ &FID $\downarrow$ &KID $\downarrow$ &$\text{FID}_c$ $\downarrow$ & $\text{KID}_c$ $\downarrow$ & \text{IS} $\uparrow$ & Time (min) $\downarrow$ \\ \midrule
SDXL-DI~\cite{sdxl}                      &  64.313   & 0.008     &  \textbf{31.042} & \textbf{0.004}   & 10.424 &  \textbf{0.648}   & 134.075      &  0.044    & \textbf{42.383}     &  \textbf{0.009} & 7.036  &   \textbf{5.456 }     \\ 
ScaleCrafter~\cite{he2023scalecrafter}                      &  67.545    & 0.013     & 60.151   & 0.020   & 11.399 &  \underline{0.653}   &  100.419     &  0.033    &  116.179    & 0.053  & 8.805  &   9.255       \\
DemoFusion~\cite{du2024demofusion}                      &  \underline{65.864}    & \underline{0.016}     &  63.001  & 0.024 & \textbf{13.282}   &  1.441    & \underline{72.378}      &    \underline{0.020}   & 94.975     & 0.045  & \underline{12.450}  &  11.382       \\ 
FouriScale~\cite{huang2024fouriscale}                      & 68.965     & 0.016     & 69.655   & 0.026  & 11.055  &  1.224     & 93.079       &   0.029    & 128.862      & 0.068  & 8.248   &   8.446      \\ 
Ours                      &  \textbf{44.723}    & \textbf{0.001}     &  \underline{36.276}   & \underline{0.006} & \underline{12.747}   &  0.853    &  \textbf{49.796}     &  \textbf{0.004}    & \underline{71.369}     &  \underline{0.029}  & \textbf{12.572}   &   \underline{6.240}     \\ \bottomrule
\end{tabular}}
\vspace{-4mm}
\end{table*}

\begin{figure*}[t]
\centering
\includegraphics[width=0.9\linewidth]{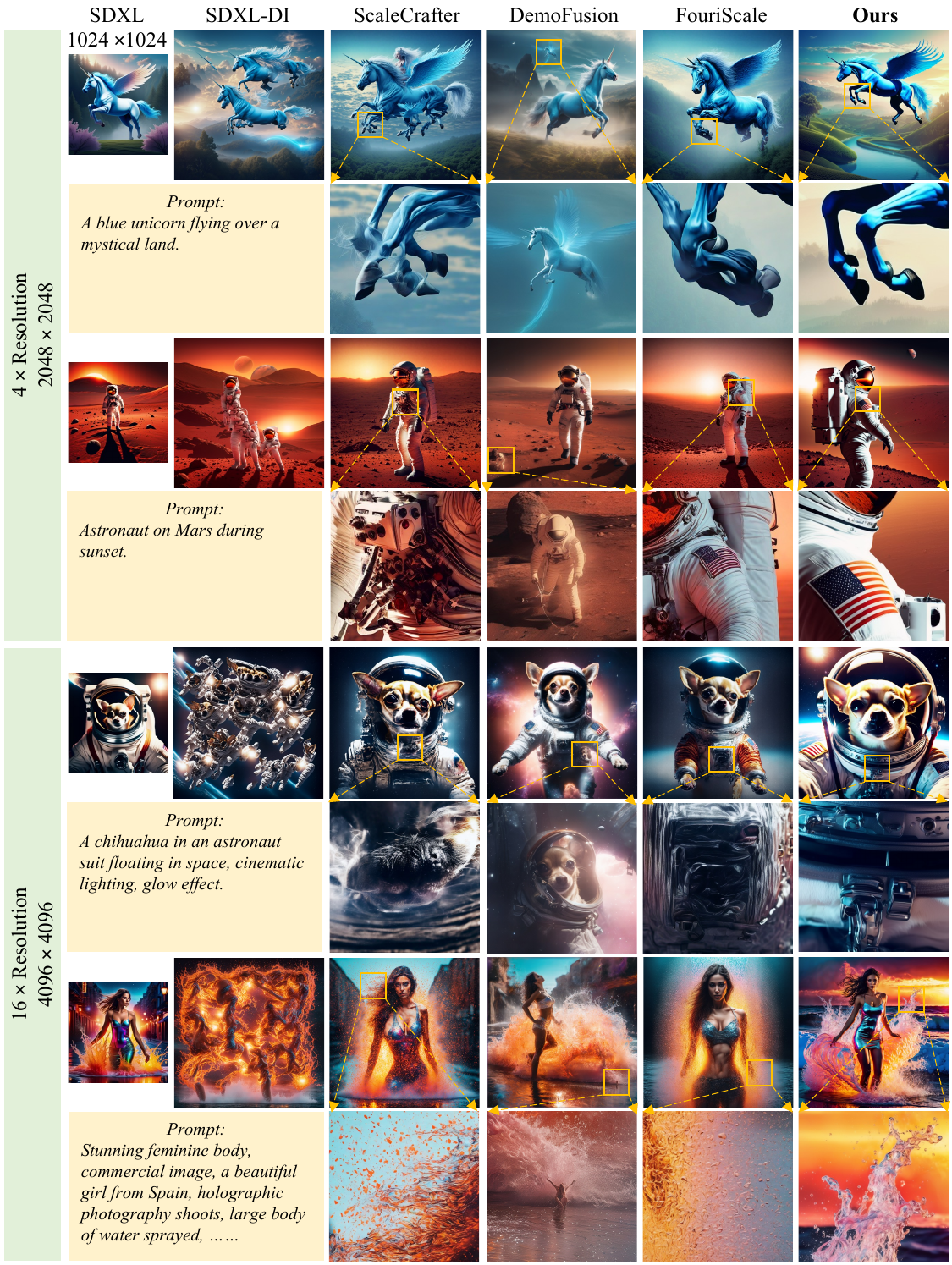}\vspace{-0.8em}
\caption{\textbf{Image qualitative comparisons with other baselines.}
Our method generates both $2048^2$ and $4096^2$ vivid images with better content coherence and local details.
Best viewed \textbf{ZOOMED-IN}.
}
\vspace{-0.8em}
\label{fig:img_comp}
\end{figure*}

\subsection{Scale Fusion}

Although tailored self-cascade upscaling and restrained dilated convolution can maintain the rough visual structures and effectively generate $4\times$ resolution images, generating $16\times$ resolution images still leads to artifacts such as local repetition, \textit{e.g.}, additional eyes or noses.
This issue arises because dilated convolution weakens the focus on local features. 
DemoFusion~\cite{du2024demofusion} addresses this by using local patches to enhance local focus. 
However, although the local patch operation mitigates local repetition, it brings small object repetition globally. 
To combine the advantages of both strategies, we design Scale Fusion, which fuses information from different receptive scales to achieve a balanced enhancement of local and global details.

Regarding global information extraction, we utilize global self-attention features. The reason is that the self-attention layer enhances the patch information based on similarity, making it easier for the subsequent cross-attention layer to aggregate semantics into a complete object. 
This can be formulated as:
\begin{equation}
\begin{aligned}
& \mathbf{h}_\text{out}^\text{global} = \text{SelfAttention}\left(\mathbf{h}_\text{in}\right) = \operatorname{softmax}\left(\frac{Q K^T}{\sqrt{d\prime}}\right) V, \\
& \text{where } Q = L_Q(\mathbf{h}_\text{in}), K = L_K(\mathbf{h}_\text{in}), V = L_V(\mathbf{h}_\text{in}).
\end{aligned}
\end{equation}
In this formulation, query $Q$, key $K$, and value $V$ are calculated from $\mathbf{h}_\text{in}$ through the linear layer $L$, and $d\prime$ is a scaling coefficient for the self-attention.

After that, the self-attention layer is independently applied to these local latent representations via $\mathbf{h}_\text{out, n} = \text{SelfAttention}\left(\mathbf{h}_\text{in, n}\right)$. And then $\mathcal{H}_\text{out}^\text{local}=\left[\mathbf{h}_\text{out, 0} \cdots, \mathbf{h}_\text{out, n} \cdots, \mathbf{h}_\text{out, N}\right]$ is reconstructed to the original size with the overlapped parts averaged as $\mathbf{h}_\text{out}^\text{local}=$ $\mathcal{R}_{\text {local}}\left(\mathcal{H}_\text{out}^\text{local}\right)$, where $\mathcal{R}_{\text {local}}$ denotes the reconstruction process.

Regarding local information extraction, we follow previous works~\cite{bar2023multidiffusion, du2024demofusion, freenoise} by calculating self-attention locally to enhance the local focus.
Specifically, we first apply a shifted crop sampling, $\mathcal{S}_{\text {local}}(\cdot)$, to obtain a series of local latent representations before each self-attention layer, \textit{i.e.}, $\mathcal{H}_\text{in}^\text{local}=\mathcal{S}_{\text {local}}\left(\mathbf{h}_\text{in}\right)=\left[\mathbf{h}_\text{in, 0} \cdots, \mathbf{h}_\text{in, n} \cdots, \mathbf{h}_\text{in, N}\right], \mathbf{h}_\text{in, n} \in \mathbb{R}^{c \times h \times w}$, where $N=\left(\frac{(H-h)}{d_h}+1\right) \times\left(\frac{(W-w)}{d_w}+1\right)$, with $d_h$ and $d_w$ representing the vertical and horizontal stride, respectively. 
After that, the self-attention layer is independently applied to these local latent representations via $\mathbf{h}_\text{out, n} = \text{SelfAttention}\left(\mathbf{h}_\text{in, n}\right)$. 
The resulting outputs $\mathcal{H}_\text{out}^\text{local}=\left[\mathbf{h}_\text{out, 0} \cdots, \mathbf{h}_\text{out, n} \cdots, \mathbf{h}_\text{out, N}\right]$ are then mapped back to the original positions, with the overlapped parts averaged to form $\mathbf{h}_\text{out}^\text{local}=$ $\mathcal{R}_{\text {local}}\left(\mathcal{H}_\text{out}^\text{local}\right)$, where $\mathcal{R}_{\text {local}}$ denotes the reconstruction process.

While $\mathbf{h}_\text{out}^\text{local}$ tends to produce better local results, it can bring unexpected small object repetition globally.
These artifacts mainly arise from dispersed high-frequency signals, which will originally be gathered to the right area through global sampling.
Therefore, we replace the high-frequency signals in the local representations with those from the global level $\mathbf{h}_\text{out}^\text{global}$:
\begin{equation}
\label{eq:fusion}
\mathbf{h}_\text{out}^\text{fusion}=\underbrace{\mathbf{h}_\text{out}^\text{global}-G\left(\mathbf{h}_\text{out}^\text{global}\right)}_{\text{high frequency}}+\underbrace{G\left(\mathbf{h}_\text{out}^\text{local}\right)}_{\text{low frequency}},
\end{equation}
where $G$ is a low-pass filter implemented as a Gaussian blur, and $\mathbf{h}_\text{out}^\text{global}-G\left(\mathbf{h}_\text{out}^\text{global}\right)$ acts as a high pass of $\mathbf{h}_\text{out}^\text{fusion}$.
 
\section{Experiments}
\label{sec:experiments}

\begin{figure*}[t]
\centering
\includegraphics[width=0.95\linewidth]{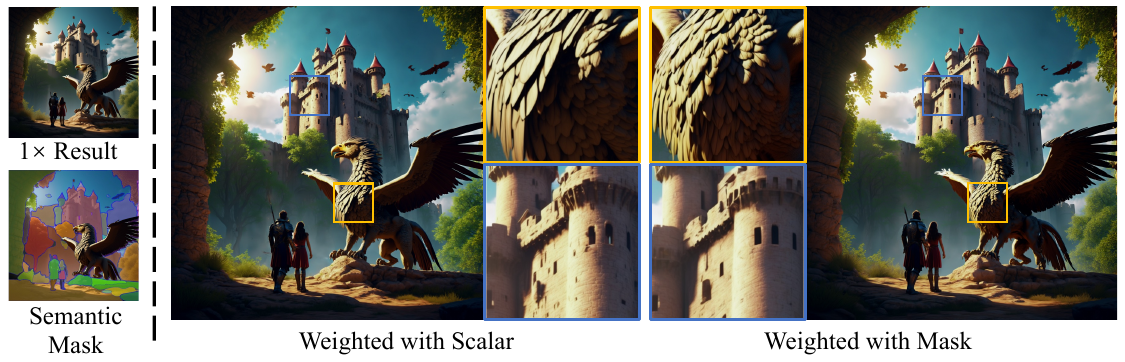}\vspace{-0.8em}
\caption{\textbf{Results of flexible control for detail level.} A better result will be generated by adding the coefficient weight in the area of Griffons and reducing the coefficient weight in the other regions. Best viewed \textbf{ZOOMED-IN}.
}
\vspace{-1.0em}
\label{fig:img_mask}
\end{figure*}

\begin{figure}[t]
\centering
\includegraphics[width=0.99\linewidth]{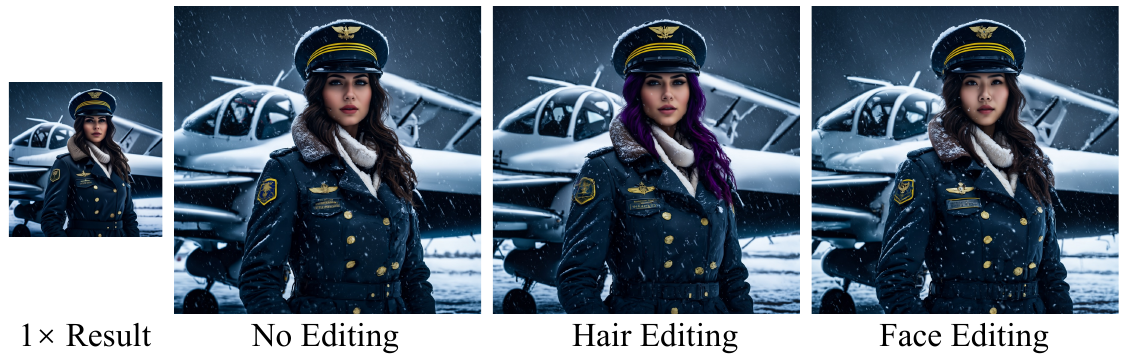}
\vspace{-0.8em}
\caption{\textbf{Results of local semantic editing.} FreeScale makes the hair purple or edits the face to make this person look more Japanese in the higher-resolution ($4096^2$). 
}
\vspace{-1.0em}
\label{fig:edit}
\end{figure}

\begin{figure}[htp]
\centering
\includegraphics[width=0.85\linewidth]{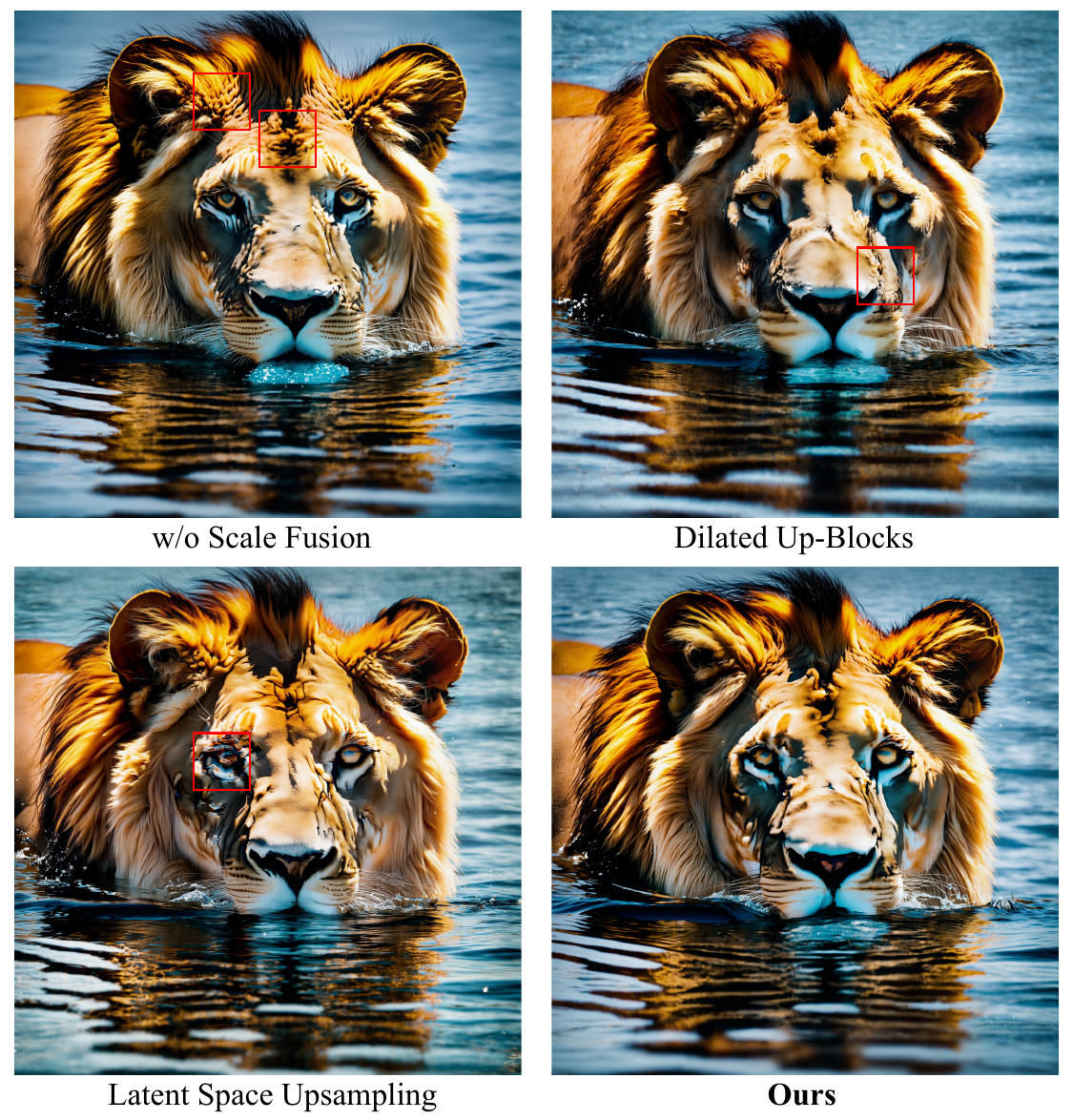}
\vspace{-0.8em}
\caption{\textbf{Qualitative image comparisons with ablations.} Our full method performs the best. The resolution of results is $4096^2$ for better visualizing the difference between the various strategies. 
}
\vspace{-1.0em}
\label{fig:img_abl}
\end{figure}

\begin{figure*}[t]
\centering
\includegraphics[width=0.99\linewidth]{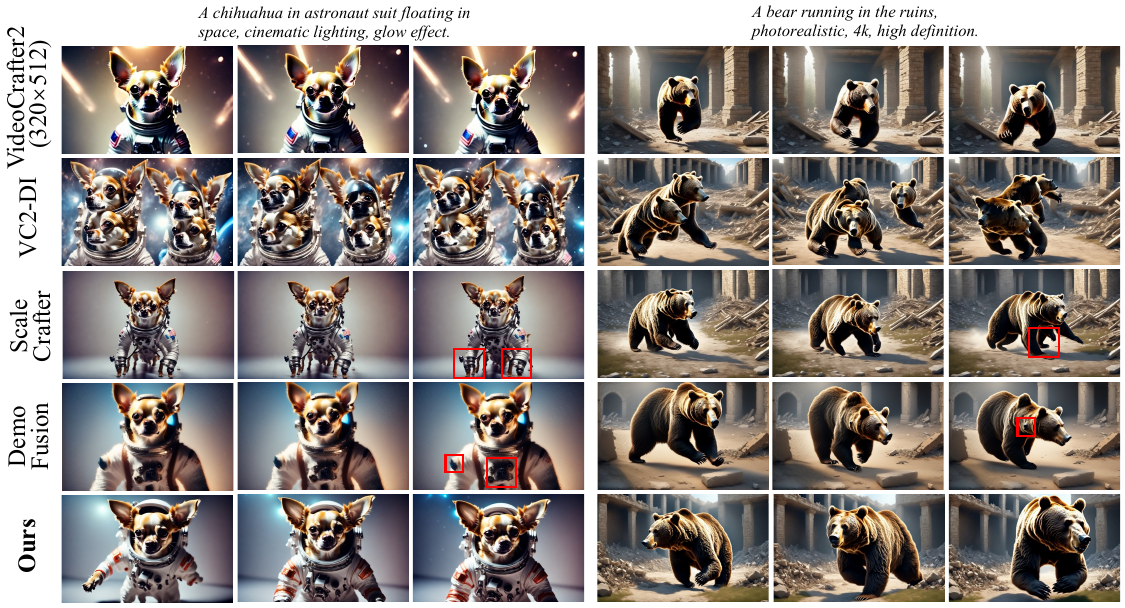}\vspace{-0.8em}
\caption{\textbf{Video qualitative comparisons with other baselines.} While other baselines fail in video generation, FreeScale effectively generates higher-resolution videos with high fidelity. Best viewed \textbf{ZOOMED-IN}.
}
\vspace{-1.0em}
\label{fig:vid_comp}
\end{figure*}

\noindent\textbf{Experimental Settings.} We conduct experiments based on an open-source T2I diffusion model SDXL~\cite{sdxl} and an open-source T2V diffusion model VideoCrafter2~\cite{chen2024videocrafter2}.
Considering the computing resources that can be afforded, we evaluate the image generation at resolutions of $2048^2$ and $4096^2$, and video generation at resolutions of $640\times1024$. All experiments are produced using a single A800 GPU.

\noindent\textbf{Evaluation Metrics.}
Since higher-resolution inference methods are intended to maintain the quality of the original resolution outputs, we calculate all metrics between the originally generated low-resolution images/videos and the corresponding high-resolution outputs.
To evaluate the quality of generated images, we report Frechet Image Distance (FID)~\cite{heusel2017gans}, Kernel Image Distance (KID) ~\cite{binkowski2018demystifying} and IS (Inception Score)~\cite{salimans2016improved}.
FID and KID need to resize the images to $299$ before the comparison and this operation may cause quality loss for high-resolution images. 
Inspired by previous work~\cite{chai2022any}, we also use cropped local patches to calculate these metrics without resizing, termed $\text{FID}_c$ and $\text{KID}_c$.
We use Frechet Video Distance (FVD)~\cite{unterthiner2018towards} to evaluate the quality of video generation. In addition, we test dynamic degree and aesthetic quality from the VBench~\cite{vbench} to evaluate the dynamics and aesthetics.

\subsection{Higher-Resolution Image Generation}

We compare FreeScale with other higher-solution image generation methods: (i) SDXL~\cite{sdxl} direct inference (SDXL-DI) (ii) ScaleCrafter~\cite{he2023scalecrafter} (iii) DemoFusion~\cite{du2024demofusion}, and (iv) FouriScale~\cite{huang2024fouriscale}. FreeU~\cite{si2023freeu} is used if compatible.

Qualitative comparison results are shown in Figure~\ref{fig:img_comp}.
We observe that direct generation often results in multiple duplicated objects and a loss of the original visual structure. 
ScaleCrafter tends to produce localized repetitions, while DemoFusion generates isolated small objects nearby. 
FouriScale can drastically alter the style for certain prompts. 
In contrast, the proposed FreeScale is capable of generating high-quality images without any unexpected repetition.

The quantitative results also confirm the superiority of FreeScale. As shown in Table~\ref{tab:comp_img}, SDXL-DI achieves the best $\text{FID}_c$ and $\text{KID}_c$. The reason is that SDXL-DI tends to generate multiple duplicated objects and its crop may be closer to the reference images. However, this behavior will sacrifice the visual structure thus SDXL gains the worst FID, KID and IS in the resolution of $4096^2$.
Overall, our approach achieves the best or second-best scores for all quality-related metrics with negligible additional time costs.

\noindent\textbf{Local Control.} FreeScale provides flexible control for detail level in generated results. Figure~\ref{fig:img_mask} shows a demo of changing the detail level of different semantic areas. During the process of tailored self-cascade upscaling, we will get $1\times$ results as intermediates. Although more details will be added or modified in the later higher-resolution stages, the overall structure and main content of the image have been determined in the $1\times$ results. It is easy to calculate semantic masks~\cite{kirillov2023segment} and assign different $\alpha$ for each region in Equation~\ref{eq:level}. As shown in Figure~\ref{fig:img_mask}, we will obtain a better result when we add the coefficient weight in the area of Griffons and reduce the coefficient weight in other regions.

In addition, this mechanism can even be extended to local semantic editing. Utilizing semantic mask from $1\times$ results, we can inject different text semantics into different regions in the layers of cross-attention. As shown in Figure~\ref{fig:edit}, FreeScale successfully edits the hair and face in the higher-resolution results.

\begin{table}[t]
\centering
\caption{\textbf{Video quantitative comparisons with baselines.} FreeScale achieves the best scores for all metrics.}
\vspace{-2mm}
\label{tab:comp_vid}
\scalebox{0.65}{\begin{tabular}{@{}l|cccc@{}}
\toprule
 \textbf{Method}  & FVD $\downarrow$  & Dynamic Degree $\uparrow$  &  Aesthetic Quality $\uparrow$ & Time (min) $\downarrow$ \\ \midrule
VC2-DI~\cite{chen2024videocrafter2} &  611.087    &  0.191    &   0.580    &  4.077     \\ 
ScaleCrafter~\cite{he2023scalecrafter} &  723.756  &  0.104    & 0.584        &  4.098     \\ 
DemoFusion~\cite{du2024demofusion} &  537.613 & 0.342     &   0.614      &  9.302      \\ 
Ours &   \textbf{484.711}   &  \textbf{0.383}   &  \textbf{0.621}     &  \textbf{3.787}     \\ \bottomrule
\end{tabular}}
\vspace{-4mm}
\end{table}

\begin{table*}[t]
\centering
\caption{\textbf{Image quantitative comparisons with other ablations.} Our final FreeScale achieves better quality-related metric scores in all experiment settings. The best results are marked in \textbf{bold}.}
\vspace{-2mm}
\label{tab:abl_img}
\scalebox{0.78}{\begin{tabular}{@{}l|cccccc|cccccc@{}}
\toprule
\multirow{2}{*}{\bf Method} & \multicolumn{6}{c|}{$2048^2$}        & \multicolumn{6}{c}{$4096^2$}        \\ \cmidrule(l){2-13} 
                        &FID $\downarrow$ &KID $\downarrow$ &$\text{FID}_c$ $\downarrow$ & $\text{KID}_c$ $\downarrow$ & \text{IS} $\uparrow$ & Time (min) $\downarrow$ &FID $\downarrow$ &KID $\downarrow$ &$\text{FID}_c$ $\downarrow$ & $\text{KID}_c$ $\downarrow$ & \text{IS} $\uparrow$ & Time (min) $\downarrow$ \\ \midrule
w/o Scale Fusion                     &  75.717     & 0.017      & 76.536    & 0.026  & 12.743   &  \textbf{0.614}        &  68.115      &  0.012     &  100.065     & 0.037   & 12.415  &  \textbf{4.566}       \\
Dilated Up-Blocks                      &  75.372     & 0.017      &  76.673   & 0.025  & 12.541   &  0.861     & 67.447       &   0.011    & 98.558      & 0.035  & 12.543   &   6.245     \\ 
Latent Space Upsampling                      & 72.454      & 0.015      & 71.793    & 0.023  & 12.210   &  0.840     & 65.081        &   0.009     & 88.632       & \textbf{0.029}  & 11.307    &   6.113     \\ 
Ours                      &  \textbf{44.723}    & \textbf{0.001}     & \textbf{36.276}   & \textbf{0.006}  & \textbf{12.747}  &  0.853     &  \textbf{49.796}     &  \textbf{0.004}    & \textbf{71.369}     &  \textbf{0.029}  & \textbf{12.572}    &  6.240     \\ \bottomrule
\end{tabular}}
\vspace{-4mm}
\end{table*}

\subsection{Higher-Resolution Video Generation}

We compare FreeScale with other tuning-free higher-solution video generation methods: (i) VideoCrafter2~\cite{chen2024videocrafter2} direct inference (VC2-DI) (ii) ScaleCrafter~\cite{he2023scalecrafter}, and (iii) DemoFusion~\cite{du2024demofusion}. FouriScale~\cite{huang2024fouriscale} is not evaluated since its bundled FreeU~\cite{si2023freeu} does not work well in video generation.

As shown in Figure~\ref{fig:vid_comp}, the behavior of VC2-DI and ScaleCrafter are similar to the corresponding version in image generation, tending to generate duplicated whole objects and local parts, respectively. However, DemoFusion has completely unexpected behavior in the video generation. Its Dilated Sampling mechanism brings strange patterns all over the frames and Skip Residual operation makes the whole video blur. In contrast, our FreeScale effectively generates higher-resolution videos with high fidelity. Table~\ref{tab:comp_vid} exhibits that our method achieves the best FVD, dynamic degree and aesthetic quality. In addition, the time cost saved by skipping certain timesteps near pure noise (transparent blocks in Figure~\ref{fig:framework}) even outweighs the extra time caused by other modules in FreeScale.

\subsection{Ablation Study}

The proposed FreeScale mainly consists of three components: (i) Tailored Self-Cascade Upscaling, (ii) Restrained Dilated Convolution, and (iii) Scale Fusion. To visually demonstrate the effectiveness of these three components, we conducted ablations on the SDXL generating $2048^2$ and $4096^2$ images. First, we show the advantage of upsampling in RGB space. As shown in Figure~\ref{fig:img_abl}, upsampling in latent space brings certain artifacts in the lion's eyes. Then dilating the convolution in up-blocks or removing Scale Fusion will cause some cluttered textures that appear in the generated results due to small repetition problems. Table~\ref{tab:abl_img} shows that our final FreeScale achieves better quality-related metric scores in all experimental settings.

\section{Conclusion}
\label{sec:conclusion}

This study introduces \textbf{FreeScale}, a tuning-free inference paradigm designed to enhance high-resolution generation capabilities in pre-trained diffusion models. By leveraging multi-scale fusion and selective frequency extraction, FreeScale effectively addresses common issues in high-resolution generation, such as repetitive patterns and quality degradation. Experimental results demonstrate the superiority of FreeScale in both image and video generation, surpassing existing methods in visual quality while also having significant advantages in inference time. FreeScale not only eliminates various forms of visual repetition but also ensures detail clarity and structural coherence in generated visuals. Additional local control capabilities provide users with more flexibility. Eventually, FreeScale achieves unprecedented \textbf{8k}-resolution text-to-image generation. 

\section{Acknowledgements}

This research is supported by the National Research Foundation, Singapore under its AI Singapore Programme (AISG Award No: AISG2-PhD-2022-01-035T), the Ministry of Education, Singapore, under its MOE AcRF Tier 2 (MOE-T2EP20221-0012, MOE-T2EP20223-0002), and Alibaba Group.
Special thanks to Yingqing He and Lanqing Guo for sharing their invaluable experience and advice in getting us started quickly in the higher-resolution visual generation task. 


{
    \small
    \bibliographystyle{ieeenat_fullname}
    \bibliography{main}
}

\appendix
\clearpage
\clearpage
\setcounter{page}{1}
\maketitlesupplementary

\noindent\textbf{Overview.} In the supplementary material, we introduce implementation details in Section~\ref{sec:implementation}, show more evaluations in Section~\ref{sec:evaluation}, exhibit more results in Section~\ref{sec:gallery}, and finally, discuss limitations and future work in Section~\ref{sec:limitation}.

\section{Implementation Details}
\label{sec:implementation}

During sampling, we perform DDIM sampling \cite{ddim} with $50$ denoising steps, setting DDIM eta to 0. For image generation, the base inference resolution of SDXL is $1024\times1024$ pixels, and the scale of the classifier-free guidance is set to $7.5$. For video generation, the base inference resolution of VideoCrafter2 is $320\times512$, the video length is $16$ frames, and the scale of the classifier-free guidance is set to $12.0$.

For tailored self-cascade upscaling, we set $K=700$ in Equation~\ref{eq:cascade} for all experiments. And in Equation~\ref{eq:level}, $\alpha$ is set as a scaler, $2$, by default. To avoid excessive and messy textures in generating 8k images, $\alpha$ is reduced to $1$. In Figure~\ref{fig:img_mask}, $\alpha$ is $3$ and $0.5$ in the targeted and other areas, respectively. Users can further adjust these parameters according to the detailed requirements of different images.
For restrained dilated convolution, the dilation factor $d$ in Equation~\ref{eq:dilation} is equal to the resolution level ($1$ represents original resolution, $2$ represents the twice height and width).
For scale fusion, the kernel size is $2\times\sqrt{height \times width \div (1024 \times 1024)} - 1$ and the standard deviation is $1$ in Equation~\ref{eq:fusion}.

\noindent\textbf{Datasets.}
We evaluate image generation on the LAION-5B dataset~\cite{schuhmann2022laion} with $1024$ randomly sampled captions.
Specifically, to better align with human preference, we randomly selected prompts from the LAION-Aesthetics-V2-6.5plus dataset to evaluate image generation. 
The LAION-Aesthetics-V2-6.5plus is a subset of the LAION 5B dataset, characterized by its high visual quality, where images have scored 6.5 or higher according to aesthetic prediction models.
Regarding the evaluation of video generation, we use randomly sampled $512$ captions from the WebVid-10M dataset~\cite{Bain21}.

\section{More Evaluation}
\label{sec:evaluation}

\begin{table}[t]
\centering
\caption{\textbf{Image quantitative comparisons with super-resolution.} Compared to super-resolution post-processing setting SDXL+Real-ESRGAN, FreeScale also achieves competitive performance. As reported in most previously published related works, higher-resolution generation methods are hard to beat SR methods completely on quantitative metrics due to the difference in difficulty between the two tasks.}
\vspace{-2mm}
\label{tab:comp_sr}
\scalebox{0.75}{\begin{tabular}{@{}l|ccccc@{}}
\toprule
 \textbf{Method}  &  FID $\downarrow$ &KID $\downarrow$ &$\text{FID}_c$ $\downarrow$ & $\text{KID}_c$ $\downarrow$ & IS $\uparrow$ \\ \midrule
SDXL+Real-ESRGAN~\cite{wang2021real} &  43.476  &  0.000 &  73.524  &  0.024  &  12.599    \\ 
Ours &   49.796     &  0.004    & 71.369     &  0.029  &  12.572    \\ \bottomrule
\end{tabular}}
\end{table}

\begin{table}[t]
\centering
\caption{\textbf{User study.} 
Users are required to pick the best one among our proposed FreeScale with the other baseline methods in terms of image-text alignment, image quality, and visual structure.}
\vspace{-2mm}
\label{tab:user}
\scalebox{0.8}{\begin{tabular}{@{}l|ccc@{}}
\toprule
 \textbf{Method}  & Text Alignment  & Image Quality  & Visual Structure \\ \midrule
SDXL-DI~\cite{sdxl}  &  0.87\%    &  0.00\%    &   0.00\%     \\ 
ScaleCrafter~\cite{he2023scalecrafter}  &  7.83\%    &  5.22\%    & 7.83\%     \\ 
DemoFusion~\cite{du2024demofusion}  &  17.39\%    &  14.35\%    &  18.26\%     \\ 
FouriScale~\cite{huang2024fouriscale}   &  2.17\%    & 2.61\%     &  1.74\%     \\ 
Ours  &  \textbf{71.74\%}    &  \textbf{77.83\%}    & \textbf{72.17\%}    \\ \bottomrule
\end{tabular}}
\end{table}

\begin{table}[th]
\centering
\caption{\textbf{User study for Video Generation.} 
Users are required to pick the best one among our proposed FreeScale with the other baseline methods in terms of text alignment, cover quality, and video quality.}
\vspace{-2mm}
\label{tab:user_vid}
\scalebox{0.9}{\begin{tabular}{@{}l|ccc@{}}
\toprule
 \textbf{Method}  & Text Alignment  & Cover Quality  & Video Quality \\ \midrule
VC2-DI  &  5.38\%    &  4.62\%    &   3.85\%     \\ 
ScaleCrafter  &  4.62\%    &  5.38\%    & 0.77\%     \\ 
DemoFusion  &  30.00\%    &  26.92\%    &  30.77\%     \\ 
Ours  &  \textbf{60.00\%}    &  \textbf{63.08\%}    & \textbf{64.62\%}    \\ \bottomrule
\end{tabular}}
\end{table}

\subsection{Comparison with Super-Resolution}

\begin{figure*}[t]
\centering
\includegraphics[width=0.99\linewidth]{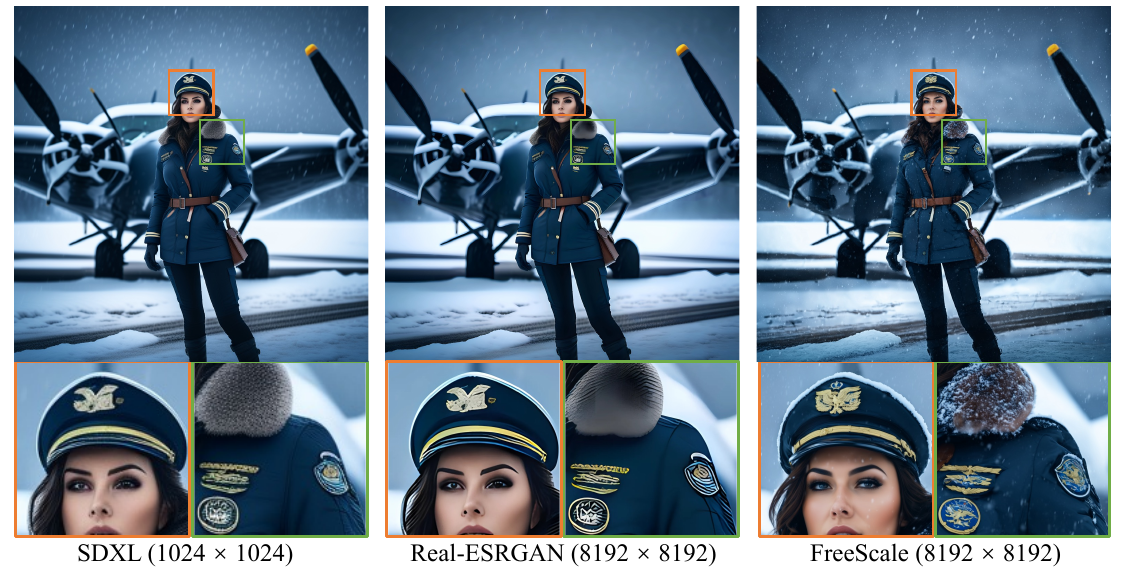}
\vspace{-0.8em}
\caption{\textbf{Image qualitative comparisons with super-resolution.} FreeScale is not inferior to SDXL+Real-ESRGAN in visual quality, and adds more details. In addition, SR methods will faithfully follow the low-resolution input while FreeScale can regenerate the original blurred areas based on the prior knowledge that the model has learned. Best viewed \textbf{ZOOMED-IN}.
}
\vspace{-1.0em}
\label{fig:diffsr}
\end{figure*}

Different from traditional super-resolution (SR) tasks. Higher-resolution generation aims to tap the potential of the pre-trained model itself. Therefore, the performance of the higher-resolution generation method is based on the base model rather than another additional SR model. We compare our method with a super-resolution post-processing setting: SDXL+Real-ESRGAN~\cite{wang2021real}. As shown in Table~\ref{tab:comp_sr}, FreeScale achieves competitive performance in quantitative metrics. As reported in most previously published related works~\cite{he2023scalecrafter, du2024demofusion}, higher-resolution generation methods are hard to beat SR methods completely on quantitative metrics due to the difference in difficulty between the two tasks. However, Figure~\ref{fig:diffsr} shows that FreeScale is not inferior to SDXL+Real-ESRGAN in visual quality, and adds more details. In addition, SR methods will faithfully follow the low-resolution input while FreeScale can regenerate the original blurred areas based on the prior knowledge that the model has learned (the eyes and logos in Figure~\ref{fig:diffsr}).

\subsection{User Study}

In addition, we conducted a user study to evaluate our results on human subjective perception. Users are asked to watch the generated images of all the methods, where each example is displayed in a random order to avoid bias, and then pick the best one in three evaluation aspects. A total of $23$ users were asked to pick the best one according to the image-text alignment, image quality, and visual structure, respectively. As shown in Table~\ref{tab:user}, our approach gains the most votes for all aspects, outperforming baseline methods by a large margin.

We also add a human study for video generation. Users were asked to pick the best one according to the text alignment, cover quality,  and video quality, respectively. As shown in Table~\ref{tab:user_vid}, our method still gains the most votes for all aspects, outperforming baseline approaches significantly.

\begin{table}[t]
\centering
\caption{\textbf{Video quantitative comparisons with other ablations.} Our final setting achieves the best or second-best scores for all metrics.
The best results are marked in \textbf{bold}, and the second best results are marked by \underline{underline}. 
}
\vspace{-2mm}
\label{tab:abl_vid}
\scalebox{0.65}{\begin{tabular}{@{}l|cccc@{}}
\toprule
 \textbf{Method}  & FVD $\downarrow$ & Dynamic Degree $\uparrow$  &  Aesthetic Quality $\uparrow$ & Time (min) $\downarrow$ \\ \midrule
Dilated Up-Blocks &  523.323    &   0.363  &  \underline{0.611}     &   \underline{3.788}      \\ 
RGB Upsampling  &  \textbf{422.245}    & \underline{0.381}   &  0.604     &   3.799   \\ 
Ours  &  \underline{484.711}   &  \textbf{0.383}   &  \textbf{0.621}    &   \textbf{3.787}    \\ \bottomrule
\end{tabular}}
\end{table}

\subsection{Ablation Study for Video Generation}

We also conduct an ablation study for higher-solution video generation. As discussed in the method part, we adopt latent space upsampling in video generation. Table~\ref{tab:abl_vid} shows that our final setting achieves the best or second-best scores for all metrics.

\section{More Results}
\label{sec:gallery}

\begin{figure}[t]
\centering
\includegraphics[width=0.99\linewidth]{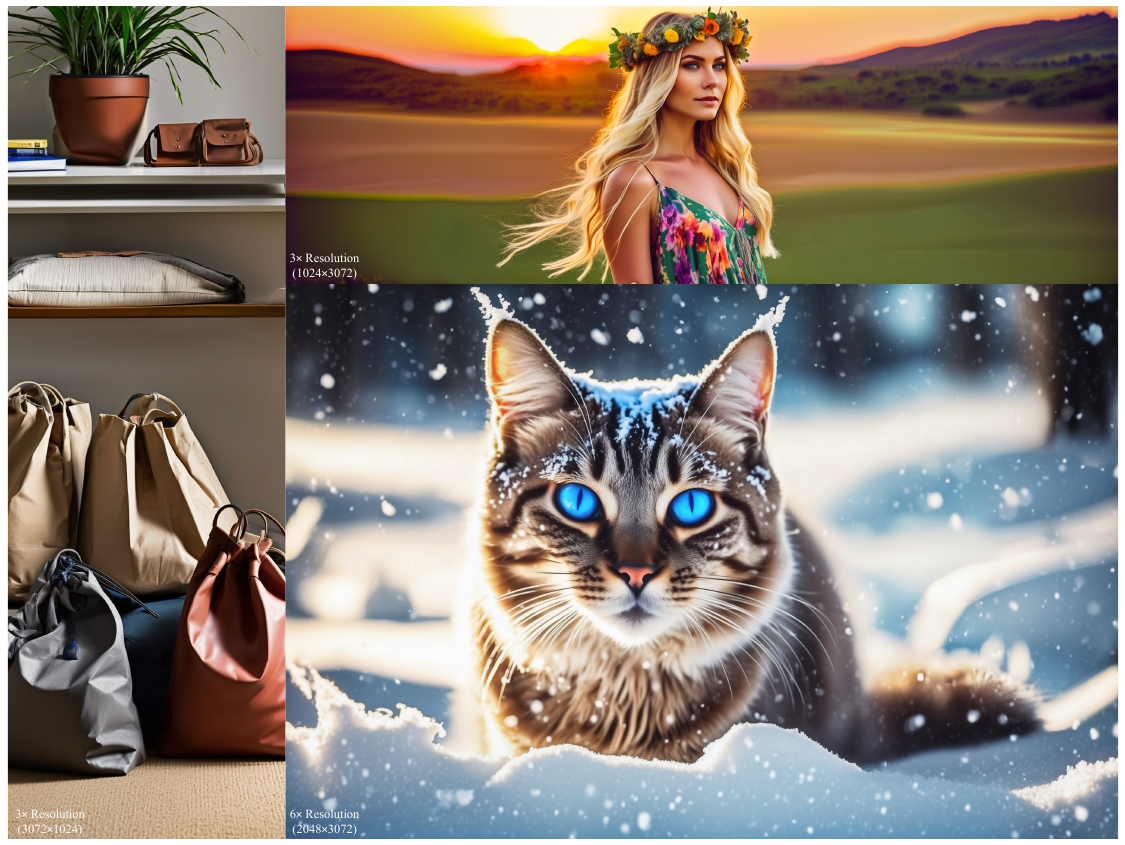}
\vspace{-0.8em}
\caption{\textbf{Flexible aspect ratio generation.} FreeScale can directly achieve a flexible aspect ratio (the resolution must be a multiple of $512$) without any adaptation.
}
\vspace{-1.0em}
\label{fig:ratio}
\end{figure}

\begin{figure*}[htb]
\centering
\includegraphics[width=0.99\linewidth]{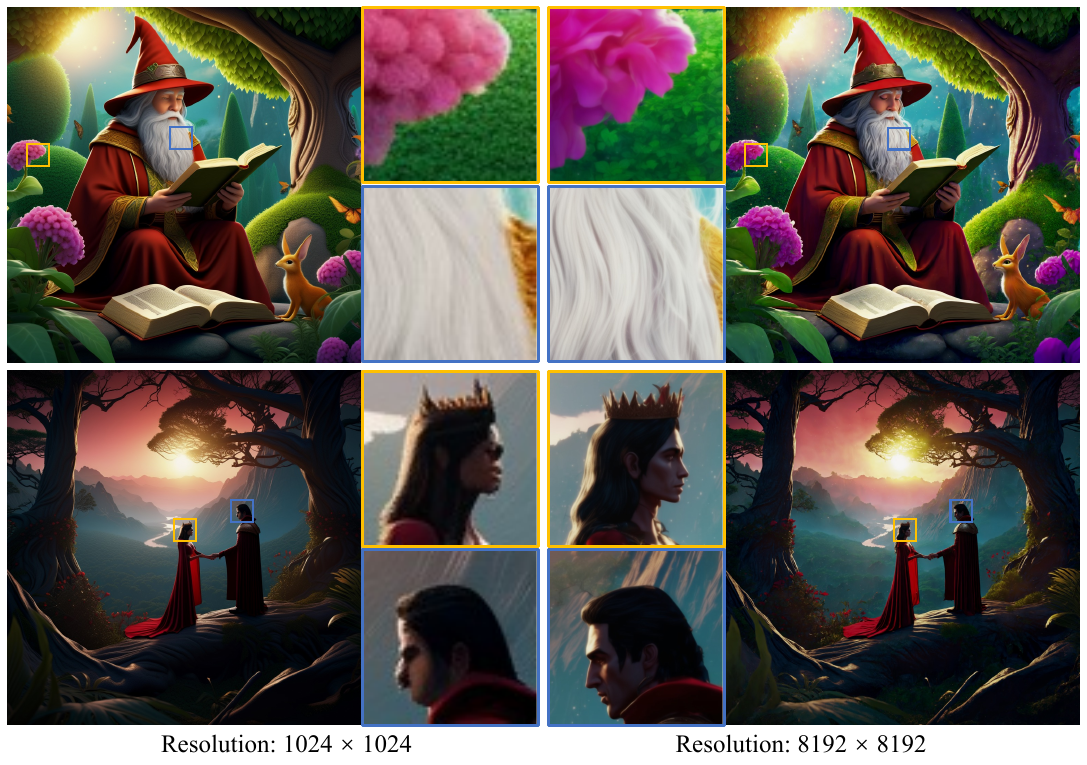}
\vspace{-0.8em}
\caption{\textbf{Zoomed in details for the 8k image.} FreeScale may regenerate the original blurred areas at low resolution based on the prior knowledge that the model has learned. As shown in the bottom row, two originally chaotic and blurry faces are clearly outlined at 8k resolution. Best viewed \textbf{ZOOMED-IN}.
}
\vspace{-1.0em}
\label{fig:img_diff8k}
\end{figure*}

\label{sec:turbo}
\begin{figure}[t]
\centering
\includegraphics[width=0.99\linewidth]{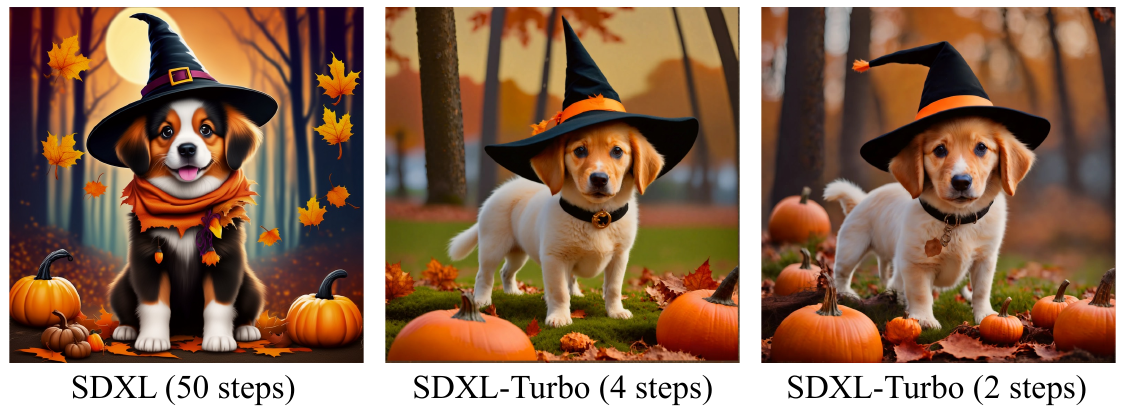}
\vspace{-0.8em}
\caption{\textbf{Fast generation with SDXL-Turbo.} FreeScale can help SDXL-Turbo generate results at $2048^2$ resolution with even $2$ timesteps.
}
\vspace{-1.0em}
\label{fig:turbo}
\end{figure}

\begin{figure*}[th]
\centering
\includegraphics[width=0.9\linewidth]{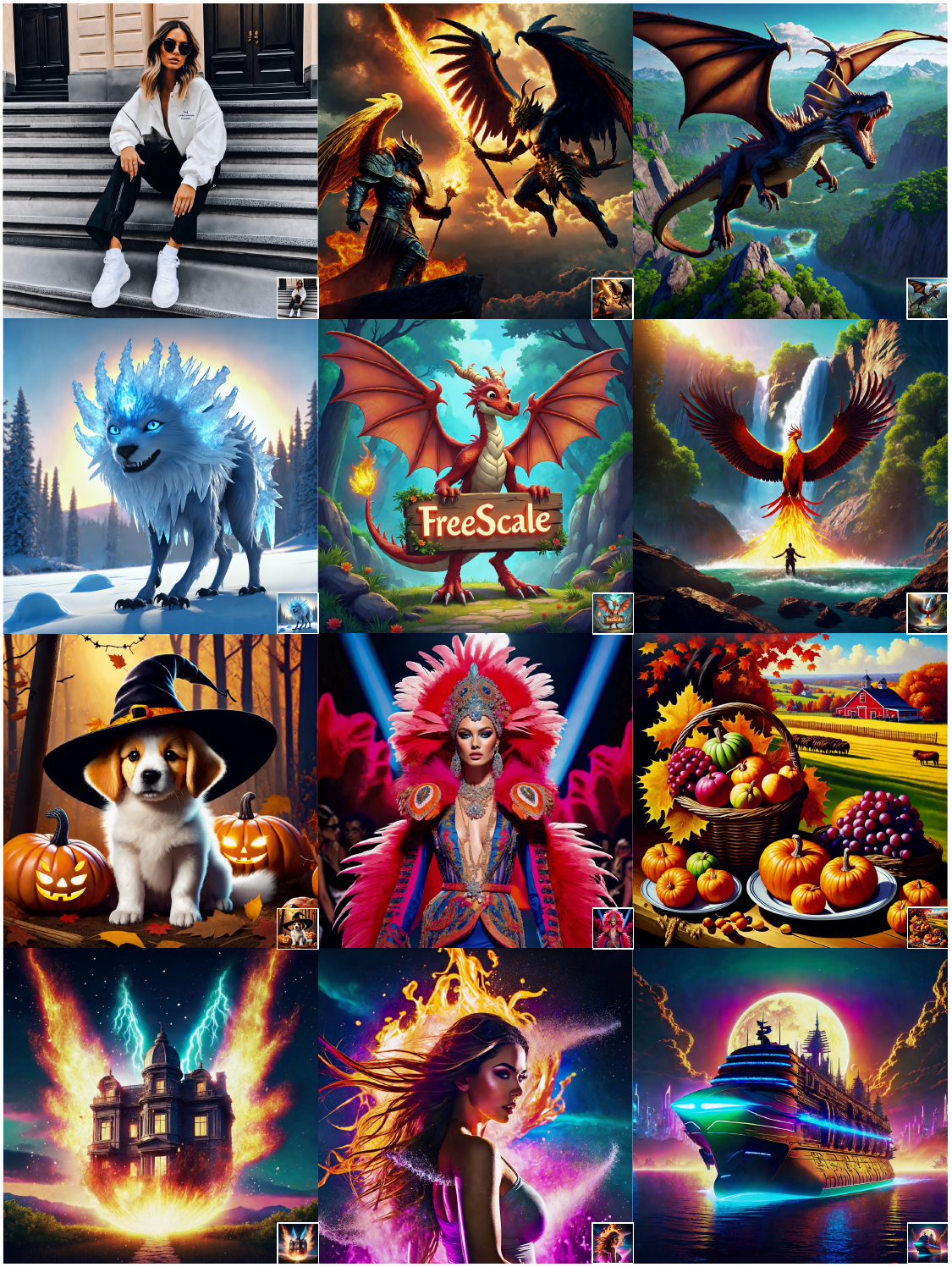}
\vspace{-0.8em}
\caption{\textbf{Gallery of generated 8k images.} We place the original-resolution result in the lower right corner for reference. FreeScale effectively enhances local details without compromising the visual structure or introducing object repetitions. Best viewed \textbf{ZOOMED-IN}.
}
\vspace{-1.0em}
\label{fig:img_all8k}
\end{figure*}

\begin{figure}[t]
\centering
\includegraphics[width=0.99\linewidth]{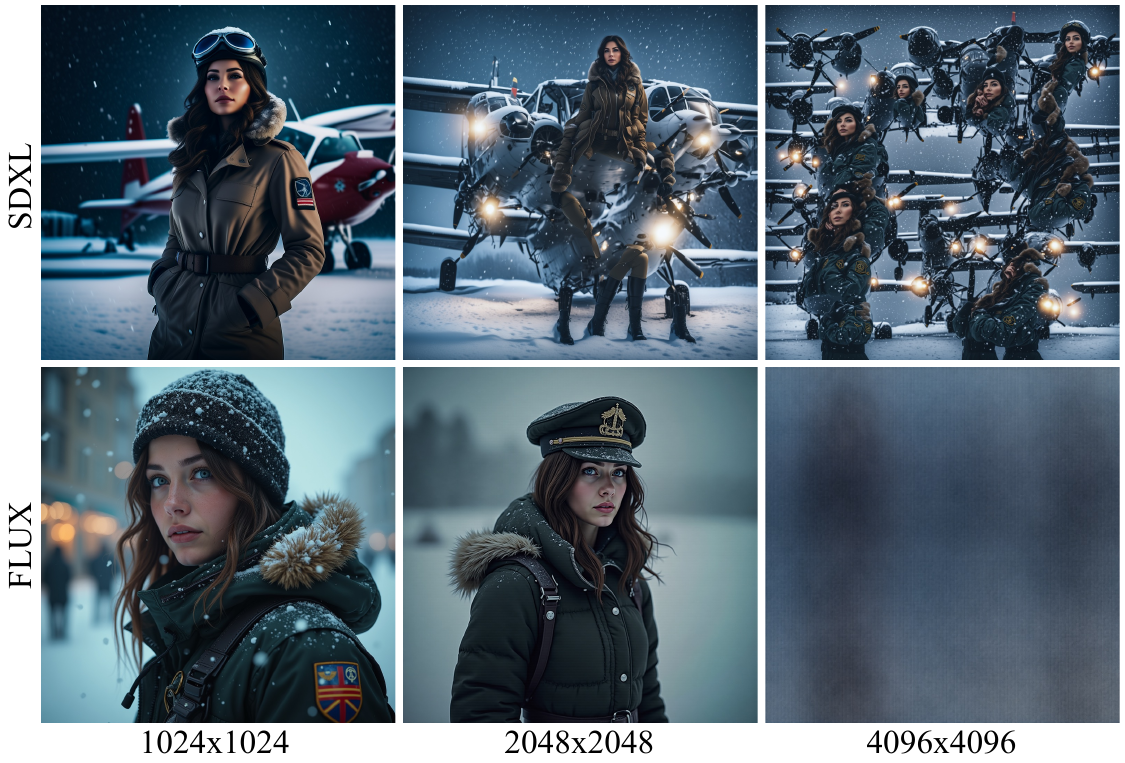}
\vspace{-0.8em}
\caption{\textbf{Structure gap.} UNet-based LDMs and DiT-based LDMs will face different challenges in the higher-resolution generation task. UNet-based LDMs face repetition problems while DiT-based LDMs face blur problems.
}
\vspace{-1.0em}
\label{fig:gap}
\end{figure}

\begin{table}[h]
\centering
\caption{\textbf{Image quantitative comparisons with baselines in $2048\times4096$ resolution.} FreeScale still achieves the best or second-best scores for all metrics.}
\vspace{-2mm}
\label{tab:comp_ratio}
\scalebox{0.95}{\begin{tabular}{@{}l|ccccc@{}}
\toprule
 \textbf{Method}  &  FID $\downarrow$ &KID $\downarrow$ &$\text{FID}_c$ $\downarrow$ & $\text{KID}_c$ $\downarrow$ & IS $\uparrow$ \\ \midrule
SDXL-DI &  97.493    &  0.026   &  \textbf{38.273}    &  \textbf{0.009}    &  7.258       \\ 
ScaleCrafter &  97.235    &  0.032   &  107.582    &  0.050    &  8.001        \\ 
DemoFusion &  \underline{72.196}    &  \underline{0.019}   &  91.264    &  0.044    &  \underline{10.622}       \\ 
FouriScale &  95.891    &  0.032   &  118.306    &  0.061    &  8.422     \\ 
Ours &   \textbf{54.704}     &  \textbf{0.004}    & \underline{65.584}     &  \underline{0.025} & \textbf{11.323}  \\ \bottomrule
\end{tabular}}
\end{table}

\subsection{Flexible Aspect Ratio Generation}

As shown in Figure~\ref{fig:ratio}, FreeScale can directly achieve a flexible aspect ratio (the resolution must be a multiple of $512$) without any adaptation. We also add quantitative experiments for $2048\times4096$ resolution. As shown in Table~\ref{tab:comp_ratio}, FreeScale still achieves the best or second-best scores for all metrics.

\subsection{Fast Generation with SDXL-Turbo}

FreeScale can easily be compatible with other models with similar structures. SDXL-Turbo~\cite{sauer2024adversarial} is a distilled version of SDXL~\cite{sdxl} and can produce similar quality results with $2\sim4$ timesteps. However, SDXL-Turbo can only generate results at $512^2$ resolution due to the knowledge loss during distillation. As shown in Figure~\ref{fig:turbo}, FreeScale can help SDXL-Turbo generate results at $2048^2$ resolution.

\subsection{Gallery of 8k Images}

Figure~\ref{fig:img_all8k} illustrates the effectiveness of FreeScale on generating ultra-high-resolution images (\ie, 8k-resolution images). 
As shown in Figure~\ref{fig:img_diff8k}, FreeScale effectively enhances local details without compromising the original visual structure or introducing object repetitions. Different from simple super-resolution, FreeScale may regenerate the original blurred areas at low resolution based on the prior knowledge that the model has learned. In Figure~\ref{fig:img_diff8k}, two originally chaotic and blurry faces are clearly outlined at 8k resolution.

\noindent\textbf{Visual Enhancement}. FreeScale also supports using existing images to replace the intermediate $1\times$ result. Compared to SDXL~\cite{sdxl}, FLUX~\cite{flux} is better in visual text generation. 
In the center of Figure~\ref{fig:img_all8k}, we first use FLUX to generate the intermediate $1\times$ result, a dragon with ``FreeScale''. 
Then we utilize the remaining pipeline of FreeScale to generate the final 8k-resolution result. 
In this sense, FreeScale is also a tool to upscale resolution and enhance detail.

\section{Limitations and Future Work}
\label{sec:limitation}

\noindent\textbf{Inference Cost.}
We employ the scale fusion only in the self-attention layers thus bringing negligible time cost. And the omitted time steps almost offset the additional cost of tailored self-cascade upscaling. As a result, the inference cost of FreeScale is close to the direct inference by the base model. However, the inference cost is still huge for ultra-high-resolution generation. 
In future work, when users require image generation at resolutions exceeding 8k, memory constraints may be mitigated through multi-GPU inference strategies, while computational efficiency can be enhanced by employing inference acceleration techniques.

\noindent\textbf{Knowledge Limitation.} Even ignoring the limitations of the computation, there is a limit to the upscaling capability of FreeScale. When the desired resolution is beyond the prior knowledge that the model has learned, no more details can be reasonably added. In other words, the endless higher-resolution result will have either the same level of detail or unnatural messy detail.
In addition, as a tuning-free framework, FreeScale’s performance relies heavily on base models. 
During the tailored self-cascade process, the intermediate $1\times$ result is equivalent to direct inference with base models. Some artifacts caused by inherently flawed (\eg, extra legs), will be inherited in further upscaling.

\noindent\textbf{Structure Gap.}
DiT-based LDMs (\eg, FLUX~\cite{flux} and CogVideoX~\cite{yang2024cogvideox}), have showcased impressive visual generation capabilities recently. 
However, UNet-based LDMs and DiT-based LDMs will face different challenges in the higher-resolution generation task. As shown in Figure~\ref{fig:gap}, UNet-based LDMs face repetition problems while DiT-based LDMs face blur problems. Most previous higher-resolution generation methods either support the UNet-based LDMs (DemoFusion~\cite{du2024demofusion}, and FouriScale~\cite{huang2024fouriscale}) or DiT-based LDMs (I-MAX~\cite{du2024max}), in line with the common sense that different problems require different strategies to solve. To support the DiT based structure, FreeScale also needs to be customized specifically.

\end{document}